# Generating Schemata of Resolution Proofs

Vincent Aravantinos, Nicolas Peltier
CNRS, LIG/TU Wien

June 2011

**Abstract**

Two distinct algorithms are presented to extract (schemata of) resolution proofs from closed tableaux for propositional schemata [4]. The first one handles the most efficient version of the tableau calculus but generates very complex derivations (denoted by rather elaborate rewrite systems). The second one has the advantage that much simpler systems can be obtained, however the considered proof procedure is less efficient.

In [2, 4] a tableau calculus (called STAB) is presented for reasoning on schemata of propositional problems. This proof procedure is able to test the validity of logical formulæ built on a set of indexed propositional symbols, using generalized connectives such as $\bigvee_{\mathtt{i}=1}^{\mathtt{n}}$ or $\bigwedge_{\mathtt{i}=1}^{\mathtt{n}}$, where $\mathtt{i},\mathtt{n}$ are part of the language ($\mathtt{n}$ denotes a parameter, i.e. an existentially quantified variable). A schema is unsatisfiable iff it is unsatisfiable for every value of $\mathtt{n}$. STAB combines the usual expansion rules of propositional logic with some delayed instantiation schemes that perform a case-analysis on the value of the parameter $\mathtt{n}$. Termination is ensured for a specific class of schemata, called *regular*, thanks to a loop detection rule which is able to prune infinite tableaux into finite ones, by encoding a form a mathematical induction (by "descente infinie"). A related algorithm, called DPLL* and based on an extension of the Davis-Putnam-Logemann-Loveland procedure, is presented in [3].

In the present work, we show that resolution proofs can be automatically extracted from the closed tableaux constructed by STAB or DPLL* on unsatisfiable schemata. More precisely, we present an algorithm that, given a closed tableau $\mathcal{T}$ for a schema $\phi_{\mathtt{n}}$, returns a schema of a refutation of $\phi_{\mathtt{n}}$ in the resolution calculus [9]. In the usual propositional case, it is well-known that algorithms exist to extract resolution proofs from closed tableaux constructed either by the usual structural rules [11, 13] or by the DPLL algorithm [7, 6]. The resolution proofs are used in various applications, for instance for certification [14], for abstraction-refinement [10] or for explanations generation [8]. The present paper extends these techniques to propositional schemata. Beside the previously mentioned applications, this turned out to be particularly important in the context of the ASAP project [1] in which schemata calculi are applied to the formalisation and analysis of mathematical proofs via cut-elimination. Indeed, the algorithm used for cut-elimination, called CERES [5], explicitly relies on the existence of a resolution proof of the so-called *characteristic clause set* extracted from the initial proof. The cut-free proof is reconstructed from this refutation, by replacing the clauses occurring in this set by some "projections" of the original proof. While STAB and DPLL* are able to detect the unsatisfiability of characteristic clause sets, as such this is completely useless since actually it is known that those sets are *always* unsatisfiable (see Proposition 3.2 in [5]). It is thus essential to be able to generate *explicitly* a representation of the resolution proof. This is precisely the aim of the present paper. Since the initial formula depends on a parameter `n`, its proof will also depend on `n` (except in very particular and trivial cases), i.e. it must be a *schema of resolution proof* (which will be encoded by recursive definitions).

The rest of the paper is structured as follows. In Section 1 we introduce the basic notions and notations used throughout our work, in particular the logic of *propositional schemata* (syntax and semantics). In Section 2 we define a tableau-based proof procedure for this logic. This calculus simulates both STAB and DPLL* (for the specific class of schemata considered in the present paper). In Section 3 we provide an algorithm to extract resolution proofs from closed tableaux. Similarly to the formulæ themselves, the constructed derivations are represented by rewrite systems. In Section 4 we introduce a second algorithm which generates simpler derivations but that requires that one of the closure rules defined in Section 2 (the so-called *Loop Detection* rule) be replaced by a less powerful rule, called the *Global Loop Detection* rule. Section 5 briefly concludes our work.

## 1 Propositional schemata

The definitions used in the present paper differ from the previous ones, but the considered logic is equivalent to the class of regular schemata considered in [2] (it is thus strictly less expressive than general schemata, for which the satisfiability problem is undecidable). We consider three disjoint sets of symbols: a set of *arithmetic variables* $\mathcal{V}$, a set of *propositional variables* $\Omega$ and a set of *defined symbols* $\Upsilon$. Let $\prec$ be a total well-founded ordering on the symbols in $\Upsilon$. An *index expression* is either a natural number or of the form $\mathtt{n} + k$, where $\mathtt{n}$ is an arithmetic variable and $k$ is a natural number. Let $I$ be a set of index expressions. The set $\mathcal{F}(I)$ of *formulæ built on* $I$ is inductively defined as follows: if $p \in \Omega \cup \Upsilon$ and $\alpha \in I$ then $p_\alpha \in \mathcal{F}(I)$; $\top, \bot \in \mathcal{F}(I)$; and if $\phi, \psi \in \mathcal{F}(I)$ then $\neg\phi$, $\phi \vee \psi$, $\phi \wedge \psi$, $\phi \Rightarrow \psi$ and $\phi \Leftrightarrow \psi$ are in $\mathcal{F}(I)$.

**Definition 1** We assume that each element $v \in \Upsilon$ is mapped to two rewrite rules $\rho_v^1$ and $\rho_v^0$ that are respectively of the form $v_{\mathtt{i}+1} \to \phi$ (inductive case) and $v_0 \to \psi$ (base case), where $\phi \in \mathcal{F}(\{\mathtt{i}+1, \mathtt{i}, 0\})$, $\psi \in \mathcal{F}(\{0\})$ and:

1. For every atom $\tau_\alpha$ occurring in $\phi$ such that $\tau \in \Upsilon$ we have either $\tau \prec v$ and $\alpha \in \{\mathtt{i}+1, \mathtt{i}, 0\}$ or $\tau = v$ and $\alpha \in \{0, \mathtt{i}\}$.

2. For every atom $\tau_\alpha$ occurring in $\psi$ such that $\tau \in \Upsilon$ we have $\tau \prec v$ and $\alpha = 0$. ◇

We denote by $\mathcal{R}$ the rewrite system: $\{\rho_v^1, \rho_v^0 \mid v \in \Upsilon\}$. The rules $\rho_v^1$ and $\rho_v^0$ are provided by the user, they encode the semantics of the defined symbols.

**Proposition 2** *$\mathcal{R}$ is convergent.*

PROOF. By Conditions 1 and 2 in Definition 1, the rules in $\mathcal{R}$ either strictly decrease the values of the defined symbols occurring in the formula w.r.t. $\prec$ or do not increase the value of these symbols but strictly decreases the value of their indices. Thus termination is obvious. Confluence is then immediate since the system is orthogonal. ∎

For every formula $\phi$, we denote by $\phi\downarrow_\mathcal{R}$ the unique normal form of $\phi$.

A *schema* (*of parameter* $\mathtt{n}$) is an element of $\mathcal{F}(\{0, \mathtt{n}, \mathtt{n}+1\})$. We denote by $\phi\{\mathtt{n} \leftarrow k\}$ the formula obtained from $\phi$ by replacing every occurrence of $\mathtt{n}$ by $k$. Obviously for any schema $\phi$, $\phi\{\mathtt{n} \leftarrow k\} \in \mathcal{F}(\{0, k, k+1\})$. A *propositional formula* is a formula $\phi \in \mathcal{F}(\mathbb{N})$ containing no defined symbols. Notice that if $\phi \in \mathcal{F}(\mathbb{N})$ then $\phi\downarrow_\mathcal{R}$ is a propositional formula.

**Proposition 3** *If $\phi \in \mathcal{F}(\mathbb{N})$ then $\phi\downarrow_\mathcal{R}$ is a propositional formula.*

PROOF. By definition of $\mathcal{R}$, $\phi\downarrow_\mathcal{R} \in \mathcal{F}(\mathbb{N})$. Furthermore, if $\phi\downarrow_\mathcal{R}$ contains a defined symbol $v$ then either $\rho_v^1$ or $\rho_v^0$ applies, which is impossible. ∎

An *interpretation* is a function mapping every arithmetic variable $\mathtt{n}$ to a natural number and every atom of the form $p_k$ (where $k \in \mathbb{N}$) to a truth value true or false. An interpretation $I$ *validates* a propositional formula $\phi$ iff one of the following conditions holds: $\phi$ is of the form $p_k$ and $I(p_k) = $ true; $\phi$ is of the form $\neg\psi$ and $I$ does not validate $\psi$; or $\phi$ is of the form $\psi_1 \vee \psi_2$ (resp. $\psi_1 \wedge \psi_2$) and $I$ validates $\psi_1$ or $\psi_2$ (resp. $\psi_1$ and $\psi_2$). $I$ *validates a schema* $\phi$ (written $I \models \phi$) iff $I$ validates $\phi\{\mathtt{n} \leftarrow I(\mathtt{n})\}\downarrow_\mathcal{R}$. We write $\phi \models \psi$ if every interpretation $I$ validating $\phi$ also validates $\psi$ and $\phi \equiv \psi$ if $\phi \models \psi$ and $\psi \models \phi$.

**Example 4** The schema $p_0 \wedge \bigwedge_{\mathtt{i}=1}^{\mathtt{n}}(p_{\mathtt{i}-1} \Rightarrow p_\mathtt{i}) \wedge \neg p_\mathtt{n}$ is encoded by $p_0 \wedge v_\mathtt{n} \wedge \neg p_\mathtt{n}$, where $v$ is defined by the rules: $v_{\mathtt{i}+1} \to (\neg p_\mathtt{i} \vee p_{\mathtt{i}+1}) \wedge v_\mathtt{i}$ and $v_0 \to \top$.

The schema $\bigvee_{\mathtt{i}=1}^{\mathtt{n}} p_\mathtt{i} \wedge \bigwedge_{\mathtt{i}=1}^{\mathtt{n}} \neg p_\mathtt{i}$ is encoded by $\tau_\mathtt{n} \wedge \tau'_\mathtt{n}$, where $\tau$ and $\tau'$ are defined by the rules: $\tau_{\mathtt{i}+1} \to p_{\mathtt{i}+1} \vee \tau_\mathtt{i}$, $\tau_0 \to \bot$, $\tau'_{\mathtt{i}+1} \to \neg p_{\mathtt{i}+1} \wedge \tau'_\mathtt{i}$ and $\tau'_0 \to \top$.

Both schemata are obviously unsatisfiable.

The schema $(p_\mathtt{n} \Leftrightarrow (p_{\mathtt{n}-1} \Leftrightarrow (\ldots (p_1 \Leftrightarrow p_0)\ldots)))$ is defined by $v'_\mathtt{n}$, where: $v'_{\mathtt{i}+1} \to (p_{\mathtt{i}+1} \Leftrightarrow v'_\mathtt{i})$ and $v'_0 \to p_0$. ♣

## 2  Proof procedure

In this section we define the proof procedure used to decide the validity of propositional schemata. We assume for simplicity that the considered schemata are in negative normal form and that the defined symbols occur only positively[1].

---
[1] If a defined symbol $v$ occurs negatively then it is easy to replace every literal of the form $\neg v_\alpha$ by an atom $\overline{v}_\alpha$ where $\overline{v}$ denotes the complementary of $v$. The rewrite rules for $\overline{v}$ are obtained by negating the right-hand side of the rules of $v$, e.g. the atom $\overline{v}$ corresponding to the symbol $v$ in Example 4 is defined by the rewrite rules $\overline{v}_{\mathtt{i}+1} \to (p_\mathtt{i} \wedge \neg p_{\mathtt{i}+1}) \vee \overline{v}_\mathtt{i}$ and $\overline{v}_0 \to \bot$.

The procedure is similar to the one presented in [2] and based on propositional block tableaux [12]. It constructs a tree labeled by finite sets of schemata, using *expansion rules* of the form: $\frac{\Phi}{\Psi_1 \mid \ldots \mid \Psi_k}$, meaning that a leaf whose label is of the form $\Phi \cup \Phi'$ (and does not already contain $\bot$) is expanded by adding $k$ children labeled by $\Phi' \cup \Psi_1, \ldots, \Phi' \cup \Psi_k$ respectively. If $\alpha$ is a node in $\mathcal{T}$, then $\mathcal{T}(\alpha)$ denotes the label of $\alpha$. The expansion rules are defined as follows:

$$\text{Normalisation:} \quad \frac{v_\alpha}{v_\alpha \downarrow_{\mathcal{R}}} \quad \text{if } v_\alpha \text{ is reducible w.r.t. } \mathcal{R}$$

$$\begin{array}{ccc}
\vee\text{-Decomposition} & \wedge\text{-Decomposition} & \text{Closure} \\
\frac{\phi \vee \psi}{\phi \mid \psi} & \frac{\phi \wedge \psi}{\phi, \psi} & \frac{\phi, \neg\phi}{\bot}
\end{array}$$

Purity rule: $\quad \frac{p_{\mathtt{n}+k}}{\top} \quad \frac{\neg p_{\mathtt{n}+k}}{\top} \quad$ if $k > 0$ and the previous rules do not apply

Note that the notion of pure literal is much simpler here than in [2]. This is due to the fact that no constant index distinct from 0 and no index of the form $\mathtt{i} + k$ where $k > 1$ are allowed.

A node that is irreducible w.r.t. all the previous rules is called a *layer*. The *Loop Detection rule* applies to nodes containing previously generated layers:

Loop Detection: $\quad \dfrac{\Phi}{\bot} \quad$ if a non leaf layer labeled by $\Phi$ exists in the tree

Note that the layer does not necessarily occur in the same branch as the one on which the rule is applied. The essential point is that the set of schemata $\Phi$ has already been considered somewhere - consequently if it has a model then an open branch necessarily exists elsewhere in the tree.

Finally, the last rule performs a case analysis on $\mathtt{n}$ (in this particular rule, $\Phi$ denotes the whole label of the considered node):

Explosion: $\quad \dfrac{\Phi}{\Phi\{\mathtt{n} \leftarrow 0\} \mid \Phi\{\mathtt{n} \leftarrow \mathtt{n} + 1\}} \quad$ if no other rule applies and $\mathtt{n}$ occurs in $\Phi$

A tableau is *closed* if the labels of all leaves contain $\bot$.

**Theorem 5** *The tableau expansion rules are terminating, i.e. there is no infinite sequence $(\mathcal{T}_i)_{i \in \mathbb{N}}$ such that for every $i \in \mathbb{N}$, $\mathcal{T}_{i+1}$ is obtained from $\mathcal{T}_i$ by applying one of the previous rules.*

PROOF. The termination of the rules Normalisation, Decomposition, Closure and Loop Detection is obvious: indeed, the Normalisation rule strictly decreases the value of the indices occurring in the formulæ, whereas the other rules cannot increase these indices and strictly reduces the size of the label (i.e. the number of symbols). Thus we only have to show that the number of layers is finite. Let

$\mathcal{S}$ be the set of layers generated by the expansion rule on a given set of schemata of some parameter $\mathtt{n}$. By definition, a layer is irreducible by the Decomposition rules, thus every formula occurring in $\mathcal{T}(\alpha)$ where $\alpha \in \mathcal{S}$ must be a literal. By irreducibility w.r.t. the Normalisation and Purity rules, the indices of these literals must be either $\mathtt{n}$ or 0. Since $\prec$ is well-founded, and since all labels are finite, the number of symbols occurring in the tableau must be finite, hence the set $\{\mathcal{T}(\alpha) \mid \alpha \in \mathcal{S}\}$ is finite. By the order of application of the expansion rules, the Explosion rule cannot be applied on two layers labeled by the same set of formulæ. Thus $\mathcal{S}$ is finite. ∎

The next theorem states that the calculus is correct:

**Theorem 6** *If $\mathcal{T}$ contains an irreducible leaf not containing $\bot$, then the label of the root of $\mathcal{T}$ is satisfiable.*

PROOF. Let $\alpha$ be the root of $\mathcal{T}$. The proof is by induction on the depth of the irreducible node in $\mathcal{T}$.

- Assume that $\alpha$ is irreducible. $\alpha$ must be a layer, thus $\mathcal{T}(\alpha)$ is a set of literals, indexed by the parameter $\mathtt{n}$ or by 0 (as shown in the proof of Theorem 5). By irreducibility w.r.t. the Explosion rule, $\mathtt{n}$ cannot occur in the label. Thus $\mathcal{T}(\alpha)$ is a set of literals indexed by 0. Furthermore, $\mathcal{T}(\alpha)$ cannot contain two complementary literals, hence must be satisfiable.

- If $\alpha$ is not irreducible, then some expansion rule must be applied on $\alpha$. The rule cannot be the Closure rule, nor the Loop Detection rule (otherwise $\alpha$ would necessarily contain $\bot$). We distinguish several cases:

    – If the $\wedge$-decomposition rule is applied on $\alpha$ then $\alpha$ has one child $\beta$. By the induction hypothesis $\mathcal{T}(\beta)$ is satisfiable. By definition of the rule, we have $\mathcal{T}(\alpha) \equiv \mathcal{T}(\beta)$ hence $\mathcal{T}(\alpha)$ is satisfiable.

    – If the $\vee$-decomposition rule is applied on $\alpha$ then $\alpha$ has two children $\beta_1$ and $\beta_2$. By definition, the irreducible node must occur in the branch of $\beta_1$ or $\beta_2$, say $\beta_1$. By the induction hypothesis $\mathcal{T}(\beta_1)$ is satisfiable. By definition of the rule, we have $\mathcal{T}(\beta_1) \models \mathcal{T}(\alpha)$ hence $\mathcal{T}(\alpha)$ is satisfiable.

    – If the Explosion rule is applied on $\alpha$ then $\alpha$ has two children $\beta_1$ and $\beta_2$ corresponding to the case $\mathtt{n} \leftarrow 0$ and $\mathtt{n} \leftarrow \mathtt{n} + 1$ respectively. By definition, the irreducible node must occur in the branch of $\beta_1$ or $\beta_2$. If it occurs in the branch of $\beta_1$, then by the induction hypothesis $\beta_1$ has a model $I$. Moreover, by definition of the rule $\mathcal{T}(\beta_1)$ contains no occurrence of $\mathtt{n}$ (since $\mathtt{n}$ is replaced by 0), thus the truth value of $\mathcal{T}(\beta_1)$ is independent of the value of $\mathtt{n}$. We may thus assume that $I(\mathtt{n}) = 0$. Then $I \models \mathcal{T}(\alpha)$ iff $I \models \mathcal{T}(\alpha)\{\mathtt{n} \leftarrow 0\}$, i.e. iff $I \models \mathcal{T}(\beta_1)$. Therefore, $I$ is a model of $\alpha$.

    If the irreducible node is a descendant of $\beta_2$, then by the induction hypothesis $\beta_2$ has a model $I$. Let $J$ be an interpretation coinciding

with $I$ except that $J(\mathtt{n}) \stackrel{\text{def}}{=} I(\mathtt{n})+1$. By definition, we have $J \models \mathcal{T}(\alpha)$ iff $I \models \mathcal{T}(\alpha)\{\mathtt{n} \leftarrow \mathtt{n}+1\}$, i.e. if $I \models \mathcal{T}(\beta_2)$. Therefore, $J$ is a model of $\alpha$.

- If the Purity rule is applied on $\alpha$, then $\alpha$ has one child $\beta$. We have $\mathcal{T}(\alpha) = \mathcal{T}(\beta) \cup \{p_{\mathtt{n}+k}\}$ (resp. $\mathcal{T}(\beta) \cup \{\neg p_{\mathtt{n}+k}\}$), where $k > 0$. By the induction hypothesis, $\mathcal{T}(\beta)$ has a model $I$. We remark that the truth value of $\mathcal{T}(\beta)$ does not depend on the value of $p_{\mathtt{n}+k}$. Indeed, $p_{\mathtt{n}+k}$ cannot occur in $\mathcal{T}(\beta)$ (otherwise the Closure rule would be applicable on $\mathcal{T}(\alpha)$ which is impossible). Furthermore, by irreducibility w.r.t. the Normalisation rule, the indices of the defined symbols occurring in $\alpha$ must be 0 or $\mathtt{n}$. Since the rewrite rules in $\mathcal{R}$ cannot increase the value of the these indices, the truth value of these indexed defined symbols depends only of the values of the atoms indexed by $I(\mathtt{n}), I(\mathtt{n}-1), \ldots, I(0)$. Thus we may assume that $I(p_{\mathtt{n}+k}) = $ true (resp. $I(p_{\mathtt{n}+k}) = $ false). Hence $I \models \mathcal{T}(\alpha)$.

∎

We will *not* prove the converse (namely that the root of every closed tableau is unsatisfiable), because this is subsumed by Theorem 18 in Section 3 (ensuring the existence of a resolution proof for every instance of the root schema).

**Example 7** The schema $\phi : p_0 \wedge \neg p_n \wedge v_n$, where $v$ is defined as in Example 4 is unsatisfiable. For instance, $\phi\{\mathtt{n} \leftarrow 2\}$ is $p_0 \wedge \neg p_2 \wedge (\neg p_0 \vee p_1) \wedge (\neg p_1 \vee p_2)$. The reader can check that the expansion rules construct the following tableau. The root is actually a layer, hence the Explosion rule is applied on it. The node (3) is deduced by the Purity rule and closed by applying the Loop Detection rule (with the root). The other rule applications are straightforward.

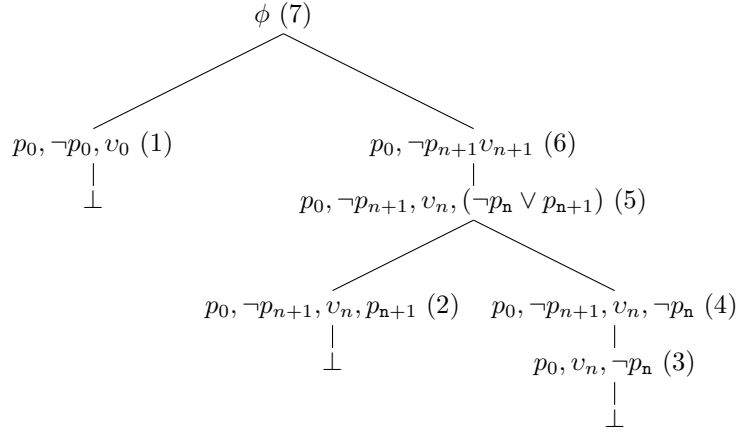

The DPLL$^*$ procedure in [3] can be simulated by the previous expansion rules, simply by adding for each propositional symbol $p \in \Omega$, a defined symbol $v^p$ with two rules: $v^p_{\mathtt{i}+1} \to ((p_{\mathtt{i}} \vee \neg p_{\mathtt{i}}) \wedge v^p_{\mathtt{i}})$ and $v^p_0 \to \top$. Then the case splitting rule of the DPLL procedure on a variable $p$ corresponds to an application of the $\wedge$-rule on $\upsilon_{n+1}^p \downarrow_\mathcal{R}$ (yielding $p_n \vee \neg p_n$) followed by an application of the $\vee$-rule on $p_n \vee \neg p_n$. The propagation rule is then simulated by combining the $\vee$-rule and the closure rule[2].

## 3 Constructing resolution proofs

### 3.1 Propositional resolution calculus

We first briefly recall the notion of resolution inference (in propositional logic). A *literal* is either an atom $p_k$ or the negation of an atom $\neg p_k$ (where $p \in \Omega$ and $k \in \mathbb{N}$). A *clause* is a (possibly empty) disjunction (or set) of literals. A *derivation* from a set of clauses $S$ is a finite sequence $C_1, \ldots, C_m$ such that for every $i \in [1, m]$, $C_i$ is either an element of $S$ or obtained from $C_1, \ldots, C_{m-1}$ by applying the resolution rule, defined as follows: $\dfrac{p_k \vee X \quad \neg p_k \vee Y}{X \vee Y}$

A *refutation* is a derivation containing $\bot$ (the empty clause). For any formula $\phi$, $\Delta$ is a *derivation from* $\phi$ if it is a derivation from a clausal form of $\phi$.

It is well-known [9] that every unsatisfiable set of (propositional) clauses has a refutation. In the context of propositional schemata, this means that every *instance* $\phi\{n \leftarrow k\}\downarrow_\mathcal{R}$ of an unsatisfiable propositional schema $\phi$ of parameter $n$ has a refutation $\Delta_k$ (which in general depends on $k$). The problem is then to construct a representation of the sequence of refutations $\Delta_0, \Delta_1, \ldots, \Delta_k, \ldots$ This sequence may be seen as a *schema of refutation* which (similarly to the semantics of the defined symbols) will be denoted by a *system of rewrite rules*. From now, we assume that the considered schema is in conjunctive normal form (i.e. it contains no conjunctions inside disjunctions, even if these conjunctions are "hidden" in the inductive definitions of the defined symbols, e.g. the schema $p_n \vee \upsilon_n$, where $\upsilon$ is defined as in Example 4 is forbidden).

### 3.2 A language for representing refutations

Additional definitions are needed to provide suitable formal languages for denoting such schemata of derivations. Let $\mathcal{D}$ and $\mathcal{X}$ be two disjoint sets of symbols (disjoint from $\mathcal{V}$, $\Omega$ and $\Upsilon$). The symbols in $\mathcal{D}$ are the $\Delta$-*symbols* and the ones in $\mathcal{X}$ are the $\Delta$-*variables*. The symbols in $\mathcal{X}$ are intended to be instantiated by schemata, whereas the symbols $d \in \mathcal{D}$ will denote schemata of refutations, defined by induction (and possibly depending on an additional argument $\Delta$ denoting a formula). We assume that $\prec$ is extended into a well-founded ordering on $\mathcal{D}$.

Formally, the set of $\Delta$-expressions is inductively defined as follows:

- All schemata and all $\Delta$-variables are $\Delta$-expressions.

---
[2]This "trick" does not actually simulate the full procedure in [3], because the latter handles schemata that are more complex than the ones considered in the present paper, possibly containing nested iterations.

- If $d \in \mathcal{D}$, $\alpha$ is an index expression and $\Delta$ is a $\Delta$-expression, then $d_\alpha$ and $d_\alpha(\Delta)$ are $\Delta$-expressions.

- If $\Delta$ and $\Gamma$ are $\Delta$-expressions then $\Delta \vee \Gamma$, $\Delta \wedge \Gamma$ and $\Delta \cdot \Gamma$ are $\Delta$-expressions.

The expression $\Delta \cdot \Gamma$ is to be interpreted as the concatenation of two sequences $\Delta$ and $\Gamma$. Note that $\Delta$-expressions can represent uniformly schemata of clauses, schemata of clause sets, or schemata of derivations (i.e. schemata of sequences of clauses). For the sake of conciseness and simplicity, the previous definition does *not* ensure that the constructions are well-typed, e.g. we can consider $\Delta$-expressions of the form $\Delta \vee \Gamma$ where $\Delta$ and $\Gamma$ are two *sequences of clauses* (which obviously does not make sense: $\Delta$ and $\Gamma$ should rather be *clauses*). But in the forthcoming definitions we will ensure that all the considered $\Delta$-expressions are well-typed.

**Example 8** Let $d \in \mathcal{D}$. Then $(p_2 \vee q_0) \cdot d_2(q_0) \cdot \neg q_0 \cdot \bot$ is a $\Delta$-expression. ♣

A $\Delta$-expression is *ground* if it contains no index variable and no $\Delta$-variable. In order to interpret (ground) $\Delta$-expressions, the value of the $\Delta$-symbols is specified using a rewrite system, exactly as schemata can be transformed into propositional formulæ by interpreting the defined symbols (using the rewrite system $\mathcal{R}$). The rewrite systems used in this section are more complicated than in the previous one, since the symbols in $\mathcal{D}$ may have an additional argument.

A $\Delta$-*substitution* is a function mapping every arithmetic variable to an index expression and every $\Delta$-variable to a $\Delta$-expression. If $\Delta$ is a $\Delta$-expression and $\sigma$ is a $\Delta$-substitution, then $\Delta\sigma$ denotes the $\Delta$-expression obtained from $\Delta$ by replacing every variable $x \in \mathcal{V} \cup \mathcal{X}$ by $\sigma(x)$.

**Definition 9** A $\mathcal{D}$-*system* is a set of *rewrite rules* of the form $\Delta \to \Gamma$, where $\Delta, \Gamma$ are two $\Delta$-expressions such that every arithmetic variable and every $\Delta$-variable occurring in $\Gamma$ also occurs in $\Delta$. A $\mathcal{D}$-system is *propositional* if it contains no $\Delta$-variables (it may contain arithmetic variables).

Given two $\Delta$-expressions $\Delta$ and $\Gamma$ and a $\mathcal{D}$-system $\mathfrak{R}$, we write $\Delta \to_{\mathfrak{R}} \Gamma$ if there exists a rule $\Delta' \to \Gamma'$ in $\mathfrak{R}$ and a $\Delta$-substitution $\sigma$ such that $\Gamma$ is obtained from $\Delta$ by replacing an occurrence of an expression $\Delta'\sigma$ by $\Gamma'\sigma$. ◇

For matching, the associativity and commutativity of logical symbols are not taken into account in general, *except for conjunctions occurring at the root level* (this rather unusual convention is needed to ensure confluence without having to bother on the order of the schemata at the root level). For instance the rule $d(p \wedge ((r \wedge q) \vee \neg r)) \to p$ does not apply on $d(p \wedge (\neg r \vee (r \wedge q)))$ nor on $d(p \wedge ((q \wedge r) \vee \neg r))$, but it applies on $d(((r \wedge q) \vee \neg r) \wedge p)$. Similarly, $d(p \wedge q) \to p$ applies on $d(p)$ by assuming $q = \top$.

**Example 10** Consider the following rewrite system ($Z$ is a $\Delta$-variable).

$$\{d_{\mathtt{i}+1}(Z) \to (\neg p_{\mathtt{i}+1} \vee p_{\mathtt{i}}) \cdot (p_{\mathtt{i}} \vee Z) \cdot d_{\mathtt{i}}(Z), \quad d_0(Z) \to \neg p_0 \cdot Z\}$$

The reader can check that it reduces the Δ-expression of Example 8 to:

$$(p_2 \vee q_0) \cdot (\neg p_2 \vee p_1) \cdot (p_1 \vee q_0) \cdot (\neg p_1 \vee p_0) \cdot (p_0 \vee q_0) \cdot \neg p_0 \cdot q_0 \cdot \neg q_0 \cdot \bot$$

This last expression is a refutation. ♣

## 3.3 From closed tableaux to resolution proofs

Let $\mathcal{T}$ be a closed tableau of a schema $\phi$. The general idea is to construct, from $\mathcal{T}$, a $\mathcal{D}$-system $\mathfrak{R}(\mathcal{T})$ representing a schema of refutation for $\phi$. Obviously, $\mathfrak{R}(\mathcal{T})$ represents an inductive proof of the assertion: "for every $\mathtt{n} \in \mathbb{N}$, the corresponding instance of $\phi$ is unsatisfiable". Ideally, we would just refute the base case, and then build a refutation of $\phi$ at $\mathtt{n}+1$ from a refutation of $\phi$ at $\mathtt{n}$. However, as often in inductive reasoning, we need to generalize the conjecture in order to refute it properly. This is done as follows: recall that our aim is to construct a refutation of $\phi$, i.e. a derivation of $\bot$ from $\phi$; instead, however, $\mathfrak{R}(\mathcal{T})$ will describe how to build a derivation of $X$ from $\phi \vee X$, *for any $X$* (formally, $X$ will be a Δ-variable). Then, our original goal will be reached by just substituting $\bot$ to $X$. In practice, we need to generalize even more this reasoning since the construction of $\mathfrak{R}(\mathcal{T})$ is done by mapping *every node $\alpha$* of $\mathcal{T}$ to some rewrite rules. So, instead of considering only the root schema $\phi$, we need to consider all the formulæ $\{\phi_1, \ldots, \phi_k\}$ that occur in $T(\alpha)$. And, instead of building a derivation of $X$ from $\phi \vee X$, we build a derivation of $X_1 \vee \cdots \vee X_k$ from $(\phi_1 \vee X_1) \wedge \cdots \wedge (\phi_k \vee X_k)$, for some Δ-variables $X_1, \ldots, X_k$. More precisely we build a derivation of a clause $C \subseteq X_1 \vee \ldots \vee X_k$, since some formulæ $\phi_i \vee X_i$ may be useless. We retrieve our original goal when we just substitute the root of $\mathcal{T}$ to $\alpha$.

The following definition constructs a $\mathcal{D}$-system $\mathfrak{R}(\mathcal{T})$ and two Δ-symbols $\nu^\alpha$ and $\mu^\alpha$ such that, if $\mathcal{T}(\alpha) = \{\phi_1, \ldots, \phi_k\}$ and $U$ denotes the formula $(\phi_1 \vee X_1) \wedge \cdots \wedge (\phi_k \vee X_k)$ then $\mu_\mathtt{n}^\alpha(U)$ denotes the above clause $C$ and $\nu_\mathtt{n}^\alpha(U)$ denotes a derivation of $C$ from $U$. This system is constructed by induction on the tableau.

**Definition 11** Let $\mathcal{T}$ be a tableau. We map every node $\alpha$ in $\mathcal{T}$ to two Δ-symbols $\nu^\alpha$ and $\mu^\alpha$. We assume that the symbols $\nu^\alpha$ and $\mu^\alpha$ are pairwise distinct. The system of rules $\mathfrak{R}(\mathcal{T})$ is defined by the rules in $\mathcal{R}$ and the following rules, for every node $\alpha$ in $\mathcal{T}$ (we distinguish several cases, according to the rule applied on $\alpha$):

- If no rule is applied on $\alpha$: $\nu_\mathtt{n}^\alpha((\bot \vee X) \wedge Y) \to X \quad \mu_\mathtt{n}^\alpha((\bot \vee X) \wedge Y) \to X$

- If the Normalisation rule is applied on $\alpha$, using a formula $\phi$, yielding a node $\beta$:
$$\nu_\mathtt{n}^\alpha((\phi \vee X) \wedge Y) \to \nu_\mathtt{n}^\beta((\phi{\downarrow}_\mathcal{R} \vee X) \wedge Y)$$
$$\mu_\mathtt{n}^\alpha((\phi \vee X) \wedge Y) \to \mu_\mathtt{n}^\beta((\phi{\downarrow}_\mathcal{R} \vee X) \wedge Y)$$

- If the Closure rule is applied on $\alpha$, using $\phi$ and $\neg\phi$:
$$\nu_\mathtt{n}^\alpha((\phi \vee X) \wedge (\neg\phi \vee Y) \wedge Z) \to (\neg\phi \vee Y) \cdot (\phi \vee X) \cdot (X \vee Y)$$

$$\mu_{\mathtt{n}}^{\alpha}((\phi \vee X) \wedge (\neg\phi \vee Y) \wedge Z) \to (X \vee Y)$$

- If $\wedge$-Decomposition is applied on $\alpha$, yielding a child $\beta$:

$$\nu_{\mathtt{n}}^{\alpha}(((\phi_1 \wedge \phi_2) \vee X) \wedge Y) \to \nu_{\mathtt{n}}^{\beta}((\phi_1 \vee X) \wedge (\phi_2 \vee X) \wedge Y)$$

$$\mu_{\mathtt{n}}^{\alpha}(((\phi_1 \wedge \phi_2) \vee X) \wedge Y) \to \mu_{\mathtt{n}}^{\beta}((\phi_1 \vee X) \wedge (\phi_2 \vee X) \wedge Y)$$

- If $\vee$-Decomposition is applied on $\alpha$ using a formula $\phi \vee \psi$ and yielding two children $\beta_1$ and $\beta_2$:

$$\nu_{\mathtt{n}}^{\alpha}(((\phi_1 \vee \phi_2) \vee X) \wedge Y) \to \nu_{\mathtt{n}}^{\beta_1}((\phi_1 \vee (\phi_2 \vee X)) \wedge Y) \cdot \nu_{\mathtt{n}}^{\beta_2}(\mu_{\mathtt{n}}^{\beta_1}((\phi_1 \vee (\phi_2 \vee X)) \wedge Y) \wedge Y)$$

$$\mu_{\mathtt{n}}^{\alpha}(((\phi_1 \vee \phi_2) \vee X) \wedge Y) \to \mu_{\mathtt{n}}^{\beta_2}(\mu_{\mathtt{n}}^{\beta_1}((\phi_1 \vee (\phi_2 \vee X)) \wedge Y) \wedge Y)$$

- If the Purity rule is applied on $\alpha$, on a formula $\phi$, yielding a node $\beta$:

$$\nu_{\mathtt{n}}^{\alpha}((\phi \vee X) \wedge Y) \to \nu_{\mathtt{n}}^{\beta}(Y) \qquad \mu_{\mathtt{n}}^{\alpha}((\phi \vee X) \wedge Y) \to \mu_{\mathtt{n}}^{\beta}(Y)$$

- If the Loop Detection rule is applied on $\alpha$, using a layer $\beta$:

$$\nu_{\mathtt{n}}^{\alpha}(X) \to \nu_{\mathtt{n}}^{\beta}(X) \qquad \mu_{\mathtt{n}}^{\alpha}(X) \to \mu_{\mathtt{n}}^{\alpha}(X)$$

- If the Explosion rule is applied on $\alpha$, yielding two children $\beta_1$ and $\beta_2$, corresponding to the cases $\mathtt{n} \leftarrow 0$ and $\mathtt{n} \leftarrow \mathtt{n}+1$ respectively:

$$\nu_{0}^{\alpha}(X) \to \nu_{0}^{\beta_1}(X) \quad \nu_{\mathtt{n}+1}^{\alpha}(X) \to \nu_{\mathtt{n}}^{\beta_1}(X) \quad \mu_{0}^{\alpha}(X) \to \mu_{0}^{\beta_2}(X) \quad \mu_{\mathtt{n}+1}^{\alpha}(X) \to \mu_{\mathtt{n}}^{\beta_2}(X)$$

$\diamond$

Note that all the symbols $\phi, \phi_1, \phi_2$ denote meta-variables, and not $\Delta$-variables (hence they cannot be instantiated during rewriting, in contrast to $X, Y, \dots$).

Before establishing the properties of $\mathfrak{R}(\mathcal{T})$, we show an example of application:

**Example 12** Consider the proof tree of Example 7. The reader can check that $\mathfrak{R}(\mathcal{T})$ contains the following rules:

$$\begin{aligned}
\nu_n^1((p_0 \vee X) \wedge (\neg p_0 \vee Y) \wedge Z) &\to (p_0 \vee X) \cdot (\neg p_0 \vee Y) \cdot (X \vee Y) \\
\mu_n^1((p_0 \vee X) \wedge (\neg p_0 \vee Y) \wedge Z) &\to X \vee Y \\
\nu_n^2((p_{n+1} \vee X) \wedge (\neg p_{n+1} \vee Y) \wedge Z) &\to (p_{n+1} \vee X) \cdot (\neg p_{n+1} \vee Y) \cdot (X \vee Y) \\
\mu_n^2((p_{n+1} \vee X) \wedge (\neg p_{n+1} \vee Y) \wedge Z) &\to X \vee Y \\
\nu_n^3(X) &\to \nu_n^7(X) \\
\mu_n^3(X) &\to \mu_n^7(X) \\
\nu_n^4((\neg p_{n+1} \vee X) \wedge Y) &\to \nu_n^3(Y) \\
\mu_n^4((\neg p_{n+1} \vee X) \wedge Y) &\to \mu_n^3(Y) \\
\nu_n^5(((\neg p_n \vee p_{n+1}) \vee X) \wedge Y) &\to \nu_n^2(((p_{n+1}) \vee (\neg p_n \vee X)) \wedge Y) \\
& \quad \cdot \nu_n^4(\mu_n^2((p_{n+1} \vee (\neg p_n \vee X)) \wedge Y) \wedge Y) \\
\mu_n^5(((\neg p_n \vee p_{n+1}) \vee X) \wedge Y) &\to \mu_n^4(\mu_n^2((p_{n+1} \vee (\neg p_n \vee X)) \wedge Y) \wedge Y) \\
\nu_n^6((v_{n+1} \vee X) \wedge Y) &\to \nu_n^5(((\neg p_n \vee p_{n+1}) \vee X) \wedge v_n \wedge Y) \\
\mu_n^6((v_{n+1} \vee X) \wedge Y) &\to \mu_n^5(((\neg p_n \vee p_{n+1}) \vee X) \wedge v_n \wedge Y) \\
\nu_0^7(X) &\to \nu_0^1(X) \\
\mu_0^7(X) &\to \mu_0^1(X) \\
\nu_{n+1}^7(X) &\to \nu_n^6(X) \\
\mu_{n+1}^7(X) &\to \mu_n^6(X)
\end{aligned}$$

The $\Delta$-expression $\nu_n^7((p_0 \vee \bot) \wedge (\neg p_n \vee \bot) \wedge (v_n \vee \bot))$ denotes a refutation of $p_0 \wedge \neg p_n \wedge v_n$. This rewrite system is complex and hardly readable, fortunately it can be simplified by instantiating the arguments when possible and by *statically* evaluating the derivations that do no depend on the value of the parameter $n$. For instance the $\Delta$-symbol $\nu_n^7$ is only called on the formula $T_n = (p_0 \vee \bot) \wedge (\neg p_n \vee \bot) \wedge (v_n \vee \bot)$. Thus the rule $\nu_n^7(X) \to \nu_0^1(X)$ may be simplified by instantiating $X$ by $T_0$ and evaluating the right-hand side: $\nu_0^7(T_0) \to p_0 \cdot \neg p_0 \cdot \bot$

Similarly, the rule $\nu_{n+1}^7(X) \to \nu_n^6(X)$ can be replaced by the following rule (in this case only a partial evaluation is possible since some parts of the derivation depend on the value of $n$): $\nu_{n+1}^7(T_{n+1}) \to (\neg p_n \vee p_{n+1}) \cdot \neg p_{n+1} \cdot \neg p_n \cdot \nu_n^7(T_n)$

The obtained system (only containing the two previous rules) is obviously much simpler than the original one, in particular it is propositional (no schema variables occur in it). To improve readability, the expression $\nu_n^7(T_n)$ could be simply replaced by a fresh symbol $\nu_n^{7'}$ (with no argument). ♣

We define the following relation $\prec_\mathcal{T}$ on the nodes in a tableau $\mathcal{T}$.

**Definition 13** Let $\mathcal{T}$ be a tableau. $\prec_\mathcal{T}$ is the least transitive relation such that $\alpha \prec_\mathcal{T} \beta$ if one of the following conditions hold:

1. Either $\alpha$ is a child of $\beta$, but $\alpha$ does not correspond to the "$n \leftarrow n + 1$" branch of an Explosion rule. This is written $\alpha \prec_\mathcal{T}^1 \beta$.

2. Or the Loop Detection rule has been applied on the node $\beta$, using the layer $\alpha$. This is written $\alpha \prec_\mathcal{T}^2 \beta$. ◇

**Proposition 14** Let $\mathcal{T}$ be a tableau. $\prec_\mathcal{T}$ is a strict partial order.

PROOF. By definition, $\mathcal{T}$ has been obtained by a sequence of application of the Expansion rules in Section 2. If $\alpha$ and $\beta$ are two non-leaf nodes in $\mathcal{T}$, we write $\alpha \triangleleft \beta$ if the expansion rule on $\alpha$ has been applied before the one of $\beta$ during this derivation, in chronological order (of course several derivations are possible, we choose one of them arbitrarily). $\triangleleft$ is obviously an ordering. Furthermore, if we have $\alpha \prec_\mathcal{T}^2 \beta$ then by the application condition of the Loop Detection rule we must have $\alpha \triangleleft \beta$, since when the rule is applied on $\beta$ the node $\alpha$ cannot be a leaf, thus an expansion rule must already have been applied on it.

Assume that $\prec_\mathcal{T}$ is not an ordering. By definition $\prec_\mathcal{T}$ is transitive, thus it must be reflexive, i.e. there is a node $\alpha$ such that $\alpha \prec_\mathcal{T} \alpha$. By definition of $\prec_\mathcal{T}$ this means that there exists a sequence of nodes $\beta_1, \ldots, \beta_k$ such that $\beta_1 = \beta_k = \alpha$ and for every $i \in [1, k-1]$, $\beta_i \prec_\mathcal{T}^\epsilon \beta_{i+1}$ (with $\epsilon = 1, 2$). If for every $i \in [1, k-1]$ we have $\beta_i \prec_\mathcal{T}^1 \beta_{i+1}$ then for all $i \in [1, k-1]$ $\beta_i$ is a child of $\beta_{i+1}$ which implies that there is a (non trivial) path in the tableau from $\alpha$ to $\alpha$. This is impossible. Thus there is at least one node $\beta_{i+1}$ such that the Loop Detection rule is applied on $\beta_{i+1}$. W.l.o.g. we can assume that $i + 1 = k$. If for every $i \in [1, k-1]$ we have $\beta_i \prec_\mathcal{T}^2 \beta_{i+1}$ we have $\beta_i \triangleleft \beta_{i+1}$, hence by transitivity $\alpha \triangleleft \alpha$, which is impossible. Let $j$ the greatest index in $[1, k-1]$ such that $\beta_j \not\prec_\mathcal{T}^2 \beta_{j+1}$. We have $\beta_j \prec_\mathcal{T}^1 \beta_{j+1} \prec_\mathcal{T}^2 \beta_{j+2} \prec_\mathcal{T}^2 \ldots \prec_\mathcal{T}^2 \beta_k$.

Since $\beta_{j+1} \prec_\mathcal{T}^2 \beta_{j+2}$, $\beta_{j+1}$ must be a layer, thus the only rule that can be applied on $\beta_{j+1}$ is the Explosion rule. Since $\beta_j \prec_\mathcal{T}^1 \beta_{j+1}$ $\beta_j$ cannot correspond to the branch $\mathtt{n} \leftarrow \mathtt{n} + 1$ of the Explosion rule. Thus it corresponds to the branch $\mathtt{n} \leftarrow 0$. But then the nodes $\beta_j, \beta_{j-1}, \ldots, \beta_1$ cannot possibly contain $\mathtt{n}$ (since no rule can introduce an occurrence of $\mathtt{n}$ in the tableau, and since by the application condition, the Loop Detection rule cannot be applied between a leaf not containing $\mathtt{n}$ and a layer containing $\mathtt{n}$). Since $\beta_1 = \beta_k$ this means that $\beta_k, \ldots, \beta_{j+1}$ contains no occurrence of $\mathtt{n}$. But in this case the Explosion rule cannot be applied on $\beta_{j+1}$, a contradiction. ∎

**Lemma 15** *Let $\mathcal{T}$ be a tableau. $\mathfrak{R}(\mathcal{T})$ is convergent.*

PROOF. We extend the ordering $\prec_\mathcal{T}$ to the $\Delta$-symbols as follows: $\nu^\alpha \prec_\mathcal{T} \nu^\beta$ and $\mu^\alpha \prec_\mathcal{T} \mu^\beta$ if $\alpha \prec_\mathcal{T} \beta$. By definition of $\prec_\mathcal{T}$, it is easy to check that all the rules above – except the $\mathtt{n}+1$-rewrite rule corresponding to the Explosion rule – strictly decrease the value of the symbols $\nu^\alpha$ and $\mu^\alpha$. Furthermore, they do not increase the value of the indices. The Explosion rule may increase the value of these symbols but strictly decreases their indices. Thus termination is obvious. Confluence is immediate: indeed, since each node is labeled by a set (and not a multiset), the system is necessarily orthogonal (note that we assume that the semantic properties of the logical connectives are not taken into account for the matching, except the AC-properties of the $\wedge$ occurring at root level). ∎

For any $\Delta$-expression $T$, we denote by $T\downarrow_{\mathfrak{R}(\mathcal{T})}$ the normal form of $T$. We now state the soundness of our algorithm.

Lemma 16 states that the rewrite system $\mathfrak{R}(\mathcal{T})$ indeed fulfils the desired property.

**Lemma 16** *Let $\mathcal{T}$ be a closed tableau. Let $\alpha$ be a node in $\mathcal{T}$. Let $k \in \mathbb{N}$. Let $\mathcal{T}(\alpha) = \{\phi_1, \ldots, \phi_n\}$. Let $X_1, \ldots, X_n$ be a set of pairwise distinct variables in $\mathcal{V}$. Let $U = (\phi_1 \vee X_1) \wedge \ldots \wedge (\phi_n \vee X_n)$. Then $\nu_k^\alpha(U){\downarrow}_{\mathfrak{R}(\mathcal{T})}$ is a derivation from $U{\downarrow}_\mathcal{R}$ of $\mu^\alpha(U){\downarrow}_{\mathfrak{R}(\mathcal{T})}$.*

PROOF. The proof is by induction on the pair $(\alpha, k)$, using the lexicographic extension of the ordering $\prec_\mathcal{T}$ on the nodes in $\mathcal{T}$ and of the usual ordering on natural numbers (this ordering is obviously well-founded since $\mathcal{T}$ is finite). We distinguish several cases, according to the expansion rule that is applied on $\alpha$.

- If no rule is applied on $\alpha$ then $\mathcal{T}(\alpha)$ must contain $\bot$. W.l.o.g., we assume that $\phi_1 = \bot$. By Definition 11, we have $\nu_k^\alpha(U){\downarrow}_{\mathfrak{R}(\mathcal{T})} = X_1$ and $\mu^\alpha(U){\downarrow}_{\mathfrak{R}(\mathcal{T})} = X_1$, thus the proof is immediate (since $X_1$ is obviously a derivation of $X_1$).

- If the closure rule is applied on $\alpha$ then $\mathcal{T}(\alpha)$ must contain two schemata $\psi$ and $\neg\psi$. W.l.o.g., we assume that $\phi_1 = \psi$ and $\phi_2 = \neg\psi$. By Definition 11, we have $\nu_k^\alpha(U){\downarrow}_{\mathfrak{R}(\mathcal{T})} = (\psi \vee X_1) \cdot (\neg\psi \vee X_2) \cdot (X_1 \vee X_2)$ and $\mu^\alpha(U){\downarrow}_{\mathfrak{R}(\mathcal{T})} = (X_1 \vee X_2)$ hence the proof is completed, since $(\psi \vee X_1) \cdot (\neg\psi \vee X_2) \cdot (X_1 \vee X_2)$ is a derivation of $X_1 \vee X_2$.

- Assume that $\wedge$-Decomposition is applied on a schema $\psi_1 \wedge \psi_2$. W.l.o.g., we assume that $\phi_1 = (\psi_1 \wedge \psi_2)$. Let $\beta$ be the child of $\alpha$. Let $U' = (\phi_2 \vee X_2) \wedge \ldots \wedge (\phi_n \vee X_n)$, i.e. we have $U = ((\psi_1 \wedge \psi_2) \vee X_1) \wedge U'$. By Definition 11, we have $\nu_k^\alpha(U){\downarrow}_{\mathfrak{R}(\mathcal{T})} = \nu_k^\beta((\psi_1 \vee X_1) \wedge (\psi_2 \wedge X_2) \wedge U'){\downarrow}_{\mathfrak{R}(\mathcal{T})}$ and $\mu^\alpha(U){\downarrow}_{\mathfrak{R}(\mathcal{T})} = \mu^\beta((\psi_1 \vee X_1) \wedge (\psi_2 \wedge X_2) \wedge U'){\downarrow}_{\mathfrak{R}(\mathcal{T})}$. Thus, by the induction hypothesis, $\nu_k^\alpha((\psi_1 \vee X_1) \wedge (\psi_2 \vee X_2) \wedge U){\downarrow}_{\mathfrak{R}(\mathcal{T})}$ is a derivation from $(\psi_1 \vee X_1) \wedge (\psi_2 \wedge X_2) \wedge U'{\downarrow}_\mathcal{R}$ of $\mu^\beta(U){\downarrow}_{\mathfrak{R}(\mathcal{T})} = \mu^\alpha(U){\downarrow}_{\mathfrak{R}(\mathcal{T})}$. Hence it is also a derivation from $U{\downarrow}_\mathcal{R}$ since $U$ and $(\psi_1 \vee X_1) \wedge (\psi_2 \wedge X_2) \wedge U'$ share the same clausal forms.

- Assume that $\vee$-Decomposition is applied on a schema $\psi_1 \vee \psi_2$. W.l.o.g., we assume that $\phi_1 = (\psi_1 \vee \psi_2)$. Let $\beta_1$ and $\beta_2$ be the children of $\alpha$ (corresponding to the schemata $\psi_1$ and $\psi_2$ respectively). Let $U' = (\phi_2 \vee X_2) \wedge \ldots \wedge (\phi_n \vee X_n)$, i.e. we have $U = ((\psi_1 \vee \psi_2) \vee X_1) \wedge U'$. By Definition 11, we have $\nu_k^\alpha(U){\downarrow}_{\mathfrak{R}(\mathcal{T})} = \nu_k^{\beta_1}(U){\downarrow}_{\mathfrak{R}(\mathcal{T})} \cdot \nu_k^{\beta_2}(\mu_k^{\beta_1}(U) \wedge U'){\downarrow}_{\mathfrak{R}(\mathcal{T})}$ and $\mu_k^\alpha(U){\downarrow}_{\mathfrak{R}(\mathcal{T})} = \mu_k^{\beta_2}(\mu_k^{\beta_1}(U) \wedge U'){\downarrow}_{\mathfrak{R}(\mathcal{T})}$.

  By the induction hypothesis, $\nu_k^{\beta_1}(U){\downarrow}_{\mathfrak{R}(\mathcal{T})}$ is a derivation from $U{\downarrow}_\mathcal{R}$ of $\mu_k^{\beta_1}(U){\downarrow}_{\mathfrak{R}(\mathcal{T})}$. Then, again by the induction hypothesis, $\nu_k^{\beta_2}(\mu_k^{\beta_1}(U) \wedge U'){\downarrow}_{\mathfrak{R}(\mathcal{T})}$ is a derivation from $\mu_k^{\beta_1}(U) \wedge U'{\downarrow}_\mathcal{R}$ of $\mu^{\beta_2}(\mu_k^{\beta_1}(U) \wedge U'){\downarrow}_{\mathfrak{R}(\mathcal{T})}$ i.e. of $\mu_k^\alpha(U){\downarrow}_{\mathfrak{R}(\mathcal{T})}$. Consequently, $\nu_k^\alpha(U){\downarrow}_{\mathfrak{R}(\mathcal{T})}$ is a derivation from $U{\downarrow}_\mathcal{R}$ of $\mu_k^\alpha(U){\downarrow}_{\mathfrak{R}(\mathcal{T})}$.

- Assume that the Loop Detection rule is applied on $\alpha$, using a node $\beta$. $\mathfrak{R}(\mathcal{T})$ contains the rule $\nu_n^\alpha(X) \rightarrow \nu_n^\beta(X)$ and $\mu_n^\alpha(X) \rightarrow \mu_n^\alpha(X)$. Then the proof is straightforward, by the induction hypothesis.

- Assume that the Purity rule is applied on $\alpha$, yielding a node $\beta$. Since $\mathcal{T}(\alpha) \supset \mathcal{T}(\beta)$, the proof is immediate (a derivation from a set $S$ is also a derivation from $S \cup S'$).

- Assume that Explosion is applied on $\alpha$, yielding two nodes $\beta_1$ and $\beta_2$ (corresponding respectively to the case $\mathtt{n} \leftarrow 0$ and $\mathtt{n} \leftarrow \mathtt{n} + 1$). We distinguish two cases, according to the value of $k$.

  - If $k = 0$ then we have $\nu_k^\alpha(U){\downarrow}_{\mathfrak{R}(\mathcal{T})} = \nu_0^{\beta_1}(U){\downarrow}_{\mathfrak{R}(\mathcal{T})}$ and $\mu_k^\alpha(U){\downarrow}_{\mathfrak{R}(\mathcal{T})} = \mu_0^{\beta_1}(U){\downarrow}_{\mathfrak{R}(\mathcal{T})}$. By the induction hypothesis, $\nu_0^{\beta_1}(U){\downarrow}_{\mathfrak{R}(\mathcal{T})}$ is a derivation from $U$ of $\mu_0^{\beta_1}(U){\downarrow}_{\mathfrak{R}(\mathcal{T})}$ hence the proof is completed.

  - If $k > 0$ then $\nu_k^\alpha(U){\downarrow}_{\mathfrak{R}(\mathcal{T})} = \nu_{k-1}^{\beta_1}(U){\downarrow}_{\mathfrak{R}(\mathcal{T})}$ and $\mu_k^\alpha(U){\downarrow}_{\mathfrak{R}(\mathcal{T})} = \mu_{k-1}^{\beta_1}(U){\downarrow}_{\mathfrak{R}(\mathcal{T})}$. By the induction hypothesis, $\nu_{k-1}^{\beta_1}(U){\downarrow}_{\mathfrak{R}(\mathcal{T})}$ is a derivation from $U$ of $\mu_{k-1}^{\beta_1}(U){\downarrow}_{\mathfrak{R}(\mathcal{T})}$ hence the proof is completed. ∎

  Note that (contrarily to all the other cases) we may have $\beta_1 \succ_\mathcal{T} \alpha$, but we are using the induction hypothesis on $\nu_{k-1}^{\beta_1}$. This is possible since $k - 1 < k$.

Furthermore, we have the following:

**Lemma 17** *Let $\mathcal{T}$ be a closed tableau. Let $\alpha$ be a node in $\mathcal{T}$. Let $\mathcal{T}(\alpha) = \{\phi_1, \ldots, \phi_n\}$. $\mu^\alpha((\phi_1 \vee X_1) \wedge \ldots \wedge (\phi_n \vee X_n)){\downarrow}_{\mathfrak{R}(\mathcal{T})}$ is a disjunction of formulæ in $X_1, \ldots, X_n$.*

PROOF. By an immediate induction on $\mu_k^\alpha$. ∎

Thus in the case in which $X_1 = \ldots X_n = \bot$, $\nu_k^\alpha(\Phi)$ denotes a refutation of $\mathcal{T}(\alpha)$, which entails the following theorem, showing the soundness of our algorithm (and entailing in particular the completeness of the tableau calculus).

**Theorem 18** *Let $\mathcal{T}$ be a closed tableau containing a node $\alpha$. Let $\mathtt{n}$ be the parameter of $\mathcal{T}(\alpha)$. Let $\mathcal{T}(\alpha) = \{\phi_1, \ldots, \phi_n\}$ and let $\Phi = (\phi_1 \vee \bot) \wedge \ldots \wedge (\phi_n \vee \bot)$.*

*For any $k \in \mathbb{N}$, $\nu_k^\alpha(\Phi\{\mathtt{n} \leftarrow k\}){\downarrow}_{\mathfrak{R}(\mathcal{T})}$ is a refutation of $\Phi\{\mathtt{n} \leftarrow k\}{\downarrow}_\mathcal{R}$. Thus $\mathcal{T}(\alpha)$ is unsatisfiable.*

PROOF. By Lemma 16, $\nu_k^\alpha(\Phi\{\mathtt{n} \leftarrow k\}){\downarrow}_{\mathfrak{R}(\mathcal{T})}$ is a derivation from $\Phi\{\mathtt{n} \leftarrow k\}{\downarrow}_\mathcal{R}$ (hence also from $\Phi\{\mathtt{n} \leftarrow k\}{\downarrow}_\mathcal{R}$) of $\mu_k^\alpha(\Phi\{\mathtt{n} \leftarrow k\}){\downarrow}_{\mathfrak{R}(\mathcal{T})}$. By Lemma 17, $\mu_k^\alpha(\Phi\{\mathtt{n} \leftarrow k\}){\downarrow}_{\mathfrak{R}(\mathcal{T})} = \bot$. ∎

Note that the size of the rewrite system $\mathfrak{R}(\mathcal{T})$ is clearly linear w.r.t. the one of the tableau $\mathcal{T}$.

The simplification phase used in Example 12 can be applied in a systematic way. However, it is not always sufficient to reduce the rewrite system into a propositional one. Actually, it is not difficult to see that as soon as a node $\alpha$ exists in the tableau on which the $\vee$-Decomposition rule is applied, yielding two branches $\beta_1$ and $\beta_2$ that are *both* looping on an ascendant of $\alpha$, then the use of schema variables cannot be avoided.

**Example 19** Consider for instance the schema: $\phi : \neg p_0 \wedge \neg q_0 \wedge (p_\mathtt{n} \vee q_\mathtt{n}) \wedge v_\mathtt{n}$, where $v$ is defined by the rules: $v_{\mathtt{i}+1} \to (q_\mathtt{i} \vee \neg p_{\mathtt{i}+1}) \wedge (p_\mathtt{i} \vee \neg q_{\mathtt{i}+1}) \wedge v_\mathtt{i}$ and $v_0 \to \top$. The following tableau is constructed:

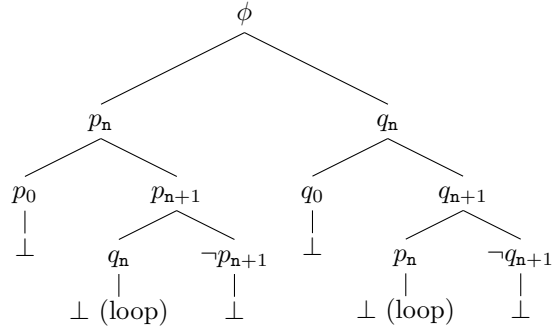

The corresponding rewrite system (after partial evaluation and simplification) is the following ($\nu_\mathtt{n}^1$ corresponds to the refutation of $\phi$):

$\nu_\mathtt{n}^1(\neg p_0 \wedge \neg q_0 \wedge (p_\mathtt{n} \vee q_\mathtt{n}) \wedge v_\mathtt{n}) \to$
$\qquad \nu_\mathtt{n}^2(\neg p_0 \wedge \neg q_0 \wedge (p_\mathtt{n} \vee q_\mathtt{n}) \wedge v_\mathtt{n}) \cdot \nu_\mathtt{n}^3(\neg p_0 \wedge \neg q_0 \wedge q_\mathtt{n} \wedge v_\mathtt{n})$
$\nu_0^2(\neg p_0 \wedge (p_0 \vee X) \wedge Y) \to \neg p_0 \cdot (p_0 \vee X) \cdot X$
$\nu_{\mathtt{n}+1}^2(\neg p_0 \wedge \neg q_0 \wedge (p_{\mathtt{n}+1} \vee X) \wedge v_{\mathtt{n}+1}) \to$
$\qquad (p_{\mathtt{n}+1} \vee X) \cdot (q_\mathtt{n} \vee \neg p_{\mathtt{n}+1}) \cdot (q_\mathtt{n} \vee X) \cdot \nu_\mathtt{n}^3(\neg p_0 \wedge \neg q_0 \wedge (q_\mathtt{n} \vee X) \wedge v_\mathtt{n})$
$\nu_0^3(\neg p_0 \wedge (q_0 \vee X) \wedge Y) \to \neg q_0 \cdot (q_0 \vee X) \cdot X$
$\nu_{\mathtt{n}+1}^3(\neg p_0 \wedge \neg q_0 \wedge (q_{\mathtt{n}+1} \vee X) \wedge v_{\mathtt{n}+1}) \to$
$\qquad (q_{\mathtt{n}+1} \vee X) \cdot (p_\mathtt{n} \vee \neg q_{\mathtt{n}+1}) \cdot (p_\mathtt{n} \vee X) \cdot \nu_\mathtt{n}^2(\neg p_0 \wedge \neg q_0 \wedge (p_\mathtt{n} \vee X) \wedge v_\mathtt{n})$

The system still contains $\Delta$-variables, although some of them have been removed by static evaluation. Note that it could be further simplified (for instance by moving the axioms such as $\neg p_0$ outside the inductive definitions), but the use of $\Delta$-variables cannot be avoided. ♣

We now focus on an alternative approach that has the advantage that only propositional rewrite systems are generated.

## 4 Globally looping tableaux

Compared to the previous approach, the second algorithm generates much simpler rewrite systems, but it has the drawback that a more restrictive version of the Loop Detection rule must be used to prune the tableaux into finite ones. At a very high and informal level: in the first approach, we were building mutually inductive proofs of several lemmata, whereas, in the second approach, we manage to have one single invariant proved by a single induction.

We first need to introduce some additional terminology. A node $\alpha$ is *of rank* $k$ in a tableau $\mathcal{T}$ of root $\beta$ if there are *exactly* $k$ applications of the Explosion rule between $\beta$ and $\alpha$ (including $\beta$, but not $\alpha$). Leaves($\mathcal{T}, \alpha$) denotes the set of non-closed leaves below $\alpha$ in $\mathcal{T}$, Layers($\mathcal{T}, k$) denotes the set of layers of rank $k$ in $\mathcal{T}$ and $\text{Layers}(\mathcal{T}, k, \alpha)$ denotes the set of layers of rank $k$ in $\mathcal{T}$ that occur below $\alpha$. For any set of formulæ $\Phi$, we denote by $\bigwedge \Phi$ the conjunction $\bigwedge_{\phi \in \Phi} \phi$. If $\mathcal{T}$ is a tableau and $N$ is a set of nodes in $\mathcal{T}$, then $\mathcal{T}[N]$ denotes the disjunction $\bigvee_{\alpha \in N} \bigwedge \mathcal{T}(\alpha)$. We write $\text{cnf}(\phi)$ for a (subsumption-minimal) clausal form of $\phi\!\downarrow_{\mathcal{R}}$.

**Definition 20** A tableau $\mathcal{T}$ is *globally looping* (w.r.t. two natural numbers $k$ and $n$) iff the following conditions hold:

1. $n < k$.

2. $\mathcal{T}[\text{Layers}(\mathcal{T}, k)] = \mathcal{T}[\text{Layers}(\mathcal{T}, n)]$ (modulo AC and idempotence).

3. All non-closed leaves in $\mathcal{T}$ are of a rank greater or equal to $k$.

Then the *Global Loop Detection rule* closes every node in $\text{Layers}(\mathcal{T}, k)$. $\diamond$

By definition, after the Global Loop Detection rule is applied, all branches containing the parameter n are closed and the construction of the tableau is over (since no leaf can be expanded anymore). Note that the Global Loop Detection rule can be simulated by several applications of the Loop Detection rule introduced in Section 2. Indeed, assume that a pair of natural numbers $(k, n)$ satisfying the conditions of Definition 20 exists. Then, by Condition 2, for every layer $\alpha$ of rank $k$, there exists a layer $\beta$ of rank $n$ such that $\mathcal{T}(\alpha) = \mathcal{T}(\beta)$. Thus the Loop Detection rule applies on $\alpha$ (w.l.o.g. we assume that the layers of rank $n$ are constructed before those of rank $k$ in all parallel branches, which is possible since $n < k$). However, it is easy to see that the converse does not hold: the Global Loop Detection rule is *strictly less general* than the looping rule. It is, however, powerful enough to ensure termination, provided that a fair strategy is used to expand the tableau, as stated by the following theorem:

**Theorem 21** *Let $(\mathcal{T}_i)_{i \in \mathbb{N}}$ be an infinite sequence of tableaux such that, for every $i \in \mathbb{N}$, $\mathcal{T}_{i+1}$ is obtained from $\mathcal{T}_i$ by applying one of the Expansion rules of Section 2, other than the Loop Detection rule. Assume, moreover, that for every $k \in \mathbb{N}$, there exists $n \in \mathbb{N}$ such that every non-closed leaf in $\mathcal{T}_n$ is of a rank greater than $k$ (i.e. no branch is indefinitely "frozen", the rank of the leaves increases indefinitely). There exists $n \in \mathbb{N}$ such that $\mathcal{T}_n$ is globally looping.*

PROOF. We have shown (see the proof of Theorem 5) that the number of sets $\mathcal{T}_i(\alpha)$ where $\alpha$ is a leaf in $\mathcal{T}_i$ is finite. Thus the set of sets of nodes $\{\mathcal{T}_i(\alpha) \mid \alpha \in \text{Layers}(\mathcal{T}_i, n)\}$ is also finite. Let $k$ be a natural number that is strictly greater than the cardinality of this set. By the hypothesis of the theorem, there exists $n \in \mathbb{N}$ such that every leaf in $\mathcal{T}_n$ is of a rank greater than $k$. By the pigeonhole argument, there exist two natural numbers $n' < k'$ such that $\mathcal{T}[\text{Layers}(\mathcal{T}_n, n')] = \mathcal{T}[\text{Layers}(\mathcal{T}_n, k')]$. Then $\mathcal{T}_n$ is globally looping. ∎

We now show that from every tableau $\mathcal{T}$, one can extract a resolution derivation from the root of $\mathcal{T}$ of the disjunction of the leaves of $\mathcal{T}$. We first restrict ourselves to tableaux built without the Explosion and Loop Detection rules. We focus on such tableaux because they correspond to the subtrees that are found "between" two layers in an tableau built without restriction on the rules. More precisely, take a layer $\alpha$ of some rank $m$ in a tableau $\mathcal{T}$ (built without restriction on the rules). Then the subtree of $\mathcal{T}$ of root $\alpha$ and whose leaves are the layers of rank $m+1$ below $\alpha$ is indeed a tree built without Explosion nor Loop Detection (by definition of a layer).

We first build derivations for such subtrees, those derivations will then be used as the base elements for building the final schema of refutation. For such a tree $\mathcal{T}$ and a node $\alpha$ of $\mathcal{T}$, the next definition introduces $\Delta(\mathcal{T}, \alpha)$, which is intended to be a derivation of $\text{cnf}(\mathcal{T}[\text{Leaves}(\mathcal{T}, \alpha)])$ from $\text{cnf}(\mathcal{T}(\alpha))$.

**Definition 22** Let $\mathcal{T}$ be a tableau constructed using the Expansion rules, *except* the Explosion and Loop Detection rules. Let $\alpha$ be a node in $\mathcal{T}$. We define a derivation $\Delta(\mathcal{T}, \alpha)$ inductively, according to the rule that is applied on $\alpha$:

- If $\alpha$ is a leaf, then $\Delta(\mathcal{T}, \alpha)$ is defined as the sequence of clauses in $\text{cnf}(\mathcal{T}(\alpha))$.

- If the Closure rule is applied on $\alpha$, using two formulæ $\phi$ and $\neg\phi$, then $\Delta(\mathcal{T}, \alpha) \stackrel{\text{def}}{=} \phi \cdot \neg\phi \cdot \bot$ (notice that since the formulæ are in NNF, $\phi$ must be an atom).

- If the Normalisation, Purity or $\wedge$-Decomposition rule is applied on $\alpha$, yielding a node $\beta$ then $\Delta(\mathcal{T}, \alpha) \stackrel{\text{def}}{=} \Delta(\mathcal{T}, \beta)$.

- Finally, assume that the $\vee$-Decomposition rule is applied on $\alpha$ yielding two nodes $\beta_1$ and $\beta_2$. Let $\Phi_1$ and $\Phi_2$ be the clausal forms of $\phi_1$ and $\phi_2$ respectively. For any $C \in \Phi_2$, let $\Lambda'(C)$ be the derivation obtained from $\Delta(\mathcal{T}, \beta_1)$ by replacing every occurrence of a clause $D \in \Phi_1$ by $D \vee C$ (and by adding the disjunction $\vee C$ to every descendant of $D$).

  For any clause $C'$ in $\text{cnf}(\mathcal{T}[\text{Leaves}(\mathcal{T}, \beta_1)])$, we construct a derivation $\Lambda''(C')$ from $\Delta(\mathcal{T}, \beta_2)$ by replacing every occurrence of a clause $D \in \Phi_2$ by $D \vee C'$ (and by adding the disjunction $\vee C'$ to every descendant of $D$). Then $\Delta(\mathcal{T}, \alpha)$ is the concatenation of all the derivations $\Lambda'(C)$ and $\Lambda''(C')$ (with $C \in \Phi_2$ and $C' \in \text{cnf}(\mathcal{T}[\text{Leaves}(\mathcal{T}, \beta_1)])$).

Only the case of disjunction is non-trivial. Informally, it does nothing more than building, for two sets of clauses $S_1$ and $S_2$, a derivation of $\text{cnf}(S_1 \vee S_2)$ from two derivations of $S_1$ and $S_2$.

The following lemma states that $\Delta(\mathcal{T}, \alpha)$ satisfies the desired property:

**Lemma 23** *Let $\mathcal{T}$ be a tableau, constructed by using the previous expansion rules, except the Explosion and Loop Detection rules. For all nodes $\alpha$ in $\mathcal{T}$, $\Delta(\mathcal{T}, \alpha)$ is a derivation of $\text{cnf}(\mathcal{T}[\text{Leaves}(\mathcal{T}, \alpha)])$ from $\text{cnf}(\mathcal{T}(\alpha))$.*

PROOF. The proof is by induction on the depth of $\mathcal{T}$. We distinguish several cases, according to the rule applied on $\alpha$.

- If $\alpha$ is a leaf then Leaves$(\mathcal{T}, \alpha) = \{\alpha\}$. Moreover, according to Definition 22, $\Delta(\mathcal{T}, \alpha)$ is the sequence of formulæ in cnf$(\mathcal{T}(\alpha))$, thus the proof is completed.

- If the Normalisation or $\wedge$-Decomposition rule is applied on $\alpha$, yielding a node $\beta$, then we have cnf$(\mathcal{T}(\alpha)) = $ cnf$(\mathcal{T}(\beta))$. Moreover, since $\alpha$ has only one child, cnf$(\mathcal{T}[\text{Leaves}(\mathcal{T}, \alpha)]) = $ cnf$(\mathcal{T}[\text{Leaves}(\mathcal{T}, \beta)])$. Hence the proof is immediate, by the induction hypothesis.

- If the Purity rule is applied on $\alpha$, using a formula $\phi$, yielding a node $\beta$, then by the induction hypothesis, $\Delta(\mathcal{T}, \beta)$ is a derivation of cnf$(\mathcal{T}[\text{Leaves}(\mathcal{T}, \beta)])$ from cnf$(\mathcal{T}(\beta))$. Since $\alpha$ has only one child, cnf$(\mathcal{T}[\text{Leaves}(\mathcal{T}, \alpha)]) = $ cnf$(\mathcal{T}[\text{Leaves}(\mathcal{T}, \beta)])$. Furthermore, cnf$(\mathcal{T}(\alpha))$ is of the form $\phi \wedge$cnf$(\mathcal{T}(\beta))$ thus $\Delta(\mathcal{T}, \beta)$ is also a derivation from cnf$(\mathcal{T}(\alpha))$. Since, by Definition 22, $\Delta(\mathcal{T}, \alpha) = \Delta(\mathcal{T}, \beta)$, the proof is completed.

- Finally, assume that the Disjunction rule is applied on $\alpha$, using a formula $\phi_1 \vee \phi_2$. This yields two nodes $\beta_1$ and $\beta_2$, corresponding respectively to $\phi_1$ and $\phi_2$. Let $\Phi_1$ and $\Phi_2$ be a cnf of $\phi_1$ and $\phi_2$ respectively. By definition, cnf$(\mathcal{T}[\text{Leaves}(\mathcal{T}, \alpha)])$ is the clausal form of the disjunction of cnf$(\mathcal{T}[\text{Leaves}(\mathcal{T}, \beta_1)])$ and cnf$(\mathcal{T}[\text{Leaves}(\mathcal{T}, \beta_2)])$, hence every clause occurring in cnf$(\mathcal{T}[\text{Leaves}(\mathcal{T}, \alpha)])$ is of the form $C_1 \vee C_2$ where $C_i$ occurs in cnf$(\mathcal{T}[\text{Leaves}(\mathcal{T}, \beta_i)])$ ($i = 1, 2$). By the induction hypothesis $\Delta(\mathcal{T}, \beta_1)$ is a derivation of cnf$(\mathcal{T}[\text{Leaves}(\mathcal{T}, \beta_1)])$ from cnf$(\mathcal{T}(\beta_1))$. Thus in particular, for every $C \in \Phi_2$, $\Lambda'(C)$ (see Definition 22 for the notations) is a derivation from cnf$(\mathcal{T}(\alpha))$ of either $C_1$ or $C_1 \vee C$. In the first case, the formula $\phi_1$ is not needed for deriving $C_1$, thus actually, $C_1$ also occurs in cnf$(\mathcal{T}[\text{Leaves}(\mathcal{T}, \beta_2)])$. Since $C_1 \vee C_2$ is subsumption-minimal, we must have actually $C_1 = C_2$ and the proof is completed. In the second case, by the induction hypothesis $\Delta(\mathcal{T}, \beta_2)$ is a derivation of $C_2$ from cnf$(\mathcal{T}(\beta_2))$, thus $\Lambda''(C_1)$ is a derivation of $C_1 \vee C_2$ from cnf$(\mathcal{T}(\alpha)) \cup \{C_1 \vee D \mid D \in \Phi_2\}$. Hence $\Delta(\mathcal{T}, \alpha)$ is a derivation of $C_1 \vee C_2$ from cnf$(\mathcal{T}(\alpha))$.

■

Thus the function $\mathcal{T}(\alpha) \to \Delta(\mathcal{T}, \alpha)$ allows us to build derivations from subtrees of a whole tableau. Intuitively, the next step is to put together those derivations according to the positions of the corresponding subtrees in the main tableau. Consider a rank $m$ in a tableau $\mathcal{T}$. One can apply the function $\Delta$ to all the (parallel) subtrees whose root is a layer of rank $m$. Then we can do the same at rank $m + 1$, append every resulting derivation to the derivation obtained from the parent tree, and go on at rank $m + 2$, etc. This intuitively gives the structure of a rewrite system where $n$ decreases each time we go to the next rank. However this gives us a tree-like structure (to every derivation corresponding to a subtree $\mathcal{U}$ we append the derivations corresponding to all the leaves of $\mathcal{U}$, and go on with the trees below those leaves) similar to the rewrite systems presented in Section 3. Instead we would like a more linear structure. So we will consider *at once* all the layers of a given rank and get only one derivation corresponding to those nodes. For this, we need a way to apply $\Delta$ to all the subtrees at once. This is actually done by building a new tableau from the subtrees.

Let $\mathcal{T}$ be a tableau of root $\alpha$. Assume that $\mathcal{T}$ is globally looping w.r.t. $n$ and $k$, with $n < k$. Let $m < k$. We denote by $\mathcal{U}(\mathcal{T}, m)$ a tableau whose root is labeled by a formula $\mathcal{T}[\text{Layers}(\mathcal{T}, m)]$ (note that we take *all the layers* of rank $m$ at a time), and obtained by applying the $\vee$ and $\wedge$-Decomposition and Closure rules (and only these rules) until irreducibility. By definition, since the root formula of $\mathcal{U}(\mathcal{T}, m)$ is the disjunction of the labels of the layers in $\text{Layers}(\mathcal{T}, m)$, every non-closed leaf $\beta$ of $\mathcal{U}(\mathcal{T}, m)$ is labeled by a set of formulæ of the form $\mathcal{T}(\gamma_\beta)$, where $\gamma_\beta \in \text{Layers}(\mathcal{T}, m)$. Furthermore, for every $\gamma \in \text{Layers}(\mathcal{T}, m)$, there exists a leaf $\beta$ of $\mathcal{U}(\mathcal{T}, m)$ such that $\gamma_\beta = \gamma$. Since $m < k$ and since by Definition 20 the leaves of $\mathcal{T}$ must be of a rank greater or equal to $k$, the node $\gamma_\beta$ cannot be a leaf of $\mathcal{T}$. This implies that some rule is applied on $\gamma_\beta$. But the only rule that is applicable on a layer (beside the Global Loop Detection rule that cannot be applied on layers of a rank distinct from $k$) is the Explosion rule. Hence $\mathcal{T}$ necessarily contains two subtableaux, written $\mathcal{T}_\beta^0$ and $\mathcal{T}_\beta^1$, of roots $\mathcal{T}(\gamma_\beta)\{\texttt{n} \leftarrow 0\}$ and $\mathcal{T}(\gamma_\beta)\{\texttt{n} \leftarrow \texttt{n} + 1\}$ respectively. Then $\mathcal{V}^0(\mathcal{T}, m)$ and $\mathcal{V}^1(\mathcal{T}, m)$ denote respectively the tableaux obtained from $\mathcal{U}(\mathcal{T}, m)\{\texttt{n} \leftarrow 0\}$ and $\mathcal{U}(\mathcal{T}, m)\{\texttt{n} \leftarrow \texttt{n} + 1\}$ by:

- Replacing every leaf $\beta$ by $\mathcal{T}_\beta^0$ and $\mathcal{T}_\beta^1$ respectively.

- Removing, in the obtained tableau, all applications of the Explosion rule[3] (and all the nodes that occur below such an application).

By definition, the leaves of $\mathcal{V}^0(\mathcal{T}, m)$ and $\mathcal{V}^1(\mathcal{T}, m)$ are layers. They correspond either to the leaves of $\mathcal{T}$ or to the nodes in $\mathcal{T}$ on which Explosion is applied (these nodes are of rank $m + 1$ in $\mathcal{T}$).

**Proposition 24** *Let $\mathcal{T}$ be a tableau that is globally looping w.r.t. two numbers $n < k$. Let $m < k$. For any non closed leaf $\beta$ of $\mathcal{U}(\mathcal{T}, m)$, $\mathcal{T}_\beta^0$ is closed and $\text{Layers}(\mathcal{T}_\beta^1, 0) = \text{Layers}(\mathcal{T}, m + 1, \gamma_\beta)$.*

PROOF. By definition, all leaves not containing $\texttt{n}$ in $\mathcal{T}$ must be closed. Thus $\mathcal{T}_\beta^0$ is closed. Furthermore, by definition, the layers of rank 0 in $\mathcal{T}_\beta^1$ are the first layers of every branch, i.e. the first layer after $\gamma_\beta$ in $\mathcal{T}$. Since $\gamma_\beta$ is a layer of rank $m$ in $\mathcal{T}$, such layers are of rank $m + 1$. ∎

**Corollary 25** *Let $\mathcal{T}$ be a tableau that is globally looping w.r.t. two numbers $n < k$. Let $m < k$. Let $\beta$ and $\beta'$ be the roots of $\mathcal{V}^0(\mathcal{T}, m)$ and $\mathcal{V}^1(\mathcal{T}, m)$ respectively. $\text{cnf}(\mathcal{T}[\text{Leaves}(\mathcal{V}^0(\mathcal{T}, m), \beta)]) = \bot$ and $\text{cnf}(\mathcal{T}[\text{Leaves}(\mathcal{V}^1(\mathcal{T}, m), \beta)]) = \text{cnf}(\mathcal{T}[\text{Layers}(\mathcal{T}, m + 1)])$.*

---
[3]Note that, although no application of the Explosion rule occurs in $\mathcal{U}(\mathcal{T}, m)$, some applications of this rule may occur in $\mathcal{T}_\beta^1$.

PROOF. The first point stems directly from Proposition 24. For the second point, we only have to remark that by definition a node occurs in Layers($\mathcal{T}, m+1$) iff it occurs in some set Layers($\mathcal{T}, m+1, \gamma_\beta$), where $\beta$ is a leaf of $\mathcal{U}(\mathcal{T}, m)$ (since the leaves of $\mathcal{U}(\mathcal{T}, m)$ are exactly the layers of rank $m$ in $\mathcal{T}$). ∎

By applying the above function $\Delta(\mathcal{T}, \alpha)$ on the two tableaux $\mathcal{V}^1(\mathcal{T}, m)$ and $\mathcal{V}^0(\mathcal{T}, m)$, we define the following derivations (where $\alpha$ denotes the root of $\mathcal{V}^1(\mathcal{T}, m)$ and $\mathcal{V}^0(\mathcal{T}, m)$):

$$\Lambda^1(\mathcal{T}, m) \stackrel{\text{def}}{=} \Delta(\mathcal{V}^1(\mathcal{T}, m), \alpha) \quad \Lambda^0(\mathcal{T}, m) \stackrel{\text{def}}{=} \Delta(\mathcal{V}^0(\mathcal{T}, m), \alpha)$$

The following lemma states essential properties of $\Lambda^1(\mathcal{T}, m)$ and $\Lambda^0(\mathcal{T}, m)$:

**Lemma 26** *Let $\mathcal{T}$ be a tableau that is globally looping w.r.t. two numbers $n < k$. Let $m < k$.*

- *$\Lambda^0(\mathcal{T}, m)$ is a refutation of cnf($\mathcal{T}$[Layers($\mathcal{T}, m$)])$\{\mathtt{n} \leftarrow 0\}$.*

- *If $m < k-1$ then $\Lambda^1(\mathcal{T}, m)$ is a derivation from cnf($\mathcal{T}$[Layers($\mathcal{T}, m$)])$\{\mathtt{n} \leftarrow \mathtt{n}+1\}$ of cnf($\mathcal{T}$[Layers($\mathcal{T}, m+1$)]).*

- *$\Lambda^1(\mathcal{T}, k-1)$ is a derivation from cnf($\mathcal{T}$[Layers($\mathcal{T}, m$)])$\{\mathtt{n} \leftarrow \mathtt{n}+1\}$ of cnf($\mathcal{T}$[Layers($\mathcal{T}, n$)]).*

PROOF. Let $\beta$ and $\beta'$ be the roots of $\mathcal{V}^1(\mathcal{T}, k)$ and $\mathcal{V}^0(\mathcal{T}, k)$ respectively. By Lemma 23, $\Lambda^1(\mathcal{T}, m)$ is a derivation from cnf($\mathcal{V}^1(\mathcal{T}, m)(\beta)$) of cnf($\mathcal{T}$[Leaves($\mathcal{V}^1(\mathcal{T}, m), \beta$)]). By definition of $\mathcal{V}^1(\mathcal{T}, m)$, the root of $\mathcal{V}^1(\mathcal{T}, m)$ is labeled by $\Phi\{\mathtt{n} \leftarrow \mathtt{n}+1\}$, where $\Phi$ is the root of $\mathcal{U}(\mathcal{T}, m)$. By definition of $\mathcal{U}(\mathcal{T}, m)$, $\Phi = \mathcal{T}$[Layers($\mathcal{T}, m$)]. Hence $\Lambda^1(\mathcal{T}, k)$ is a derivation from cnf($\mathcal{T}$[Layers($\mathcal{T}, m$)])$\{\mathtt{n} \leftarrow \mathtt{n}+1\}$. Similarly, $\Lambda^0(\mathcal{T}, m)$ is a derivation from cnf($\mathcal{T}$[Layers($\mathcal{T}, m$)])$\{\mathtt{n} \leftarrow 0\}$.

By Corollary 25, cnf($\mathcal{T}$[Leaves($\mathcal{V}^1(\mathcal{T}, m), \beta$)]) = cnf($\mathcal{T}$[Layers($\mathcal{T}, m+1$)]). Furthermore, if $m = k-1$, then since $\mathcal{T}$ is globally looping we have cnf($\mathcal{T}$[Layers($\mathcal{T}, m+1$)]) = cnf($\mathcal{T}$[Layers($\mathcal{T}, n$)]).

Similarly, cnf($\mathcal{T}$[Leaves($\mathcal{V}^1(\mathcal{T}, m), \beta$)]) = $\bot$. ∎

Let $\mathcal{T}$ be a tableau that is globally looping w.r.t. two numbers $n < k$. We associate to each natural number $m < k$ a symbol $\gamma^m$. Let $\mathfrak{R}^\star(\mathcal{T})$ the system containing the following rules. Note that $\mathcal{V}^0(\mathcal{T}, m)$ and $\mathcal{V}^1(\mathcal{T}, m)$ are defined only w.r.t. the rank $m$, but not w.r.t. a particular node. Thus, contrarily to the transformation of Section 3, there is not one derivation per node, but rather one derivation per rank.

$$\gamma_0^m \to \Lambda^0(\mathcal{T}, m) \quad \gamma_{\mathtt{n}+1}^m \to \Lambda^1(\mathcal{T}, m) \cdot \gamma_{\mathtt{n}}^{m+1} \text{ (if } m+1 < k\text{)} \quad \gamma_{\mathtt{n}+1}^{k-1} \to \Lambda^1(\mathcal{T}, k) \cdot \gamma_{\mathtt{n}}^{n}$$

Intuitively, we are appending the derivations, rank after rank, until we reach the rank $k$ where the Global Loop Detection applies. In this case we get back at the rank of looping $n$. Thus we can see the use of grouping the derivations by rank (instead of node) as it allows to benefit from the simplified form of looping induced by the Global Loop Detection rule. In the end, the resulting rewrite system is indeed much simpler.

**Proposition 27** $\mathfrak{R}^\star(\mathcal{T})$ *is convergent.*

PROOF. Termination is easy to obtain since the rules in $\mathfrak{R}^\star(\mathcal{T})$ strictly decreases the value of the indices of the symbols $\gamma^k$. Furthermore, $\mathfrak{R}^\star(\mathcal{T})$ is obviously orthogonal. ∎

Note that, by definition, $\mathfrak{R}^\star(\mathcal{T})$ is always propositional (unlike $\mathfrak{R}(\mathcal{T})$).

**Theorem 28** *Let $\mathcal{T}$ be a tableau of root $\alpha$ that is globally looping w.r.t. two numbers $n, k$, with $n < k$. Let $m < k$. For all $i \in \mathbb{N}$, $\gamma_i^m \downarrow_{\mathfrak{R}^\star(\mathcal{T})}$ is a refutation of $cnf(\mathcal{T}[Layers(\mathcal{T}, m)])\{\mathtt{n} \leftarrow i\}\downarrow_\mathcal{R}$. Thus in particular, if $\alpha$ is a layer, $\gamma_i^0 \downarrow_{\mathfrak{R}^\star(\mathcal{T})}$ is a refutation of $\mathcal{T}(\alpha)\{\mathtt{n} \leftarrow i\}\downarrow_\mathcal{R}$.*

PROOF. This follows by induction on $i$. If $i = 0$ then we have, by definition of the rules in $\mathfrak{R}^\star(\mathcal{T})$: $\gamma_i^m \downarrow_{\mathfrak{R}^\star(\mathcal{T})} = \Lambda^0(\mathcal{T}, m)\downarrow_\mathcal{R}$. By Lemma 26 (first point), $\Lambda^0(\mathcal{T}, m)\downarrow_\mathcal{R}$ is a refutation of $\mathrm{cnf}(\mathcal{T}[\mathrm{Layers}(\mathcal{T}, m)])\{\mathtt{n} \leftarrow 0\}\downarrow_\mathcal{R}$.

If $i > 0$ then we have $\gamma_i^m \downarrow_{\mathfrak{R}^\star(\mathcal{T})} = \Lambda^1(\mathcal{T}, m)\downarrow_\mathcal{R} \{\mathtt{n} \leftarrow i\} \cdot \gamma_{i-1}^{m+1} \downarrow_{\mathfrak{R}^\star(\mathcal{T})}$. If $m < k - 1$, then by Lemma 26 (second point), $\Lambda^1(\mathcal{T}, m)$ is a derivation from $\mathrm{cnf}(\mathcal{T}[\mathrm{Layers}(\mathcal{T}, m)])\{\mathtt{n} \leftarrow \mathtt{n} + 1\}$ of $\mathrm{cnf}(\mathcal{T}[\mathrm{Layers}(\mathcal{T}, m+1)])$, hence $\Lambda^1(\mathcal{T}, i-1) \downarrow_\mathcal{R}$ is a derivation from $\mathrm{cnf}(\mathcal{T}[\mathrm{Layers}(\mathcal{T}, m)])\{\mathtt{n} \leftarrow i\} \downarrow_\mathcal{R}$ of $\mathrm{cnf}(\mathcal{T}[\mathrm{Layers}(\mathcal{T}, m+1)])\{\mathtt{n} \leftarrow i-1\}\downarrow_\mathcal{R}$. Then by the induction hypothesis, $\gamma_{i-1}^{m+1}\downarrow_{\mathfrak{R}^\star(\mathcal{T})}$ is a refutation of $\mathrm{cnf}(\mathcal{T}[\mathrm{Layers}(\mathcal{T}, m+1)])\alpha\{\mathtt{n} \leftarrow i-1\}\downarrow_\mathcal{R}$.

If $m = k-1$, then by Lemma 26 (second point), $\Lambda^1(\mathcal{T}, m)$ is a derivation from $\mathrm{cnf}(\mathcal{T}[\mathrm{Layers}(\mathcal{T}, m)])\{\mathtt{n} \leftarrow \mathtt{n}+1\}$ of $\mathrm{cnf}(\mathcal{T}[\mathrm{Layers}(\mathcal{T}, n)])$, hence $\Lambda^1(\mathcal{T}, i-1)\downarrow_\mathcal{R}$ is a derivation from $\mathrm{cnf}(\mathcal{T}[\mathrm{Layers}(\mathcal{T}, m)])\{\mathtt{n} \leftarrow i\}\downarrow_\mathcal{R}$ of $\mathrm{cnf}(\mathcal{T}[\mathrm{Layers}(\mathcal{T}, n)])\{\mathtt{n} \leftarrow i-1\}\downarrow_\mathcal{R}$. Then by the induction hypothesis, $\gamma_{i-1}^n \downarrow_{\mathfrak{R}^\star(\mathcal{T})}$ is a refutation of $\mathrm{cnf}(\mathcal{T}[\mathrm{Layers}(\mathcal{T}, n)])\alpha\{\mathtt{n} \leftarrow i-1\}\downarrow_\mathcal{R}$.

When $\alpha$ is not a layer, the rewrite system is easily adapted by prepending the derivation obtained by applying $\Delta$ to the subtree of $\mathcal{T}$ whose leaves are the layers of rank 0.

**Example 29** Consider the tableau of Example 19. This tableau is actually globally looping. The following rewrite system is constructed (after partial evaluation and simplification):

$$\gamma_0 \quad \to \quad p_0 \vee q_0 \cdot \neg p_0 \cdot q_0 \cdot \neg q_0 \cdot \bot$$
$$\gamma_{\mathtt{n}+1} \quad \to \quad (p_{\mathtt{n}+1} \vee q_{\mathtt{n}+1}) \cdot (q_{\mathtt{n}} \vee \neg p_{\mathtt{n}+1}) \cdot (q_{\mathtt{n}} \vee q_{\mathtt{n}+1}) \cdot (p_{\mathtt{n}} \vee \neg q_{\mathtt{n}+1}) \cdot (q_{\mathtt{n}} \vee p_{\mathtt{n}}) \cdot \gamma_{\mathtt{n}}$$

Compared with the system produced by the previous method (see Example 19), these rules are obviously simpler (no schema variable are needed, and only linear recursion is used). Furthermore, it is easy to check that they generate much shorter derivations. ♣

# 5 Conclusion

Two distinct algorithms have been designed for extracting schemata of resolution proofs from closed tableaux. This work is motivated by the fact that such refutations are needed for some natural applications of schemata calculus (unsatisfiability detection is not always sufficient). In particular, the explicit generation of the proofs (even in the form of proof schemata) makes possible the certification of the results produced by the provers. The first algorithm tackles the tableau calculus in its full generality, but it yields very complex representations of the derivations (which will make them less usable in practice, in particular they are not very informative for a human user). The second one uses a less powerful calculus, but it generates schemata of refutations in a much simpler format (propositional rewrite systems are obtained).

There is thus a natural trade-off between the two presented methods: none of them is uniformly superior to the other. The choice between the two algorithms should be made according to the considered applications, and/or to the form of the constructed tableaux. In some cases, as shown by the examples in Section 3, the first approach generates a propositional rewrite system. In this case it should of course be preferred. Future work includes the implementation of the two methods and the precise evaluation of the complexity of the second algorithm. One could also wonder whether a polynomial algorithm generating propositional derivations exists for the general case. We conjecture that the use of $\Delta$-variables cannot be avoided in general.